\documentclass[runningheads]{llncs}

\usepackage{eccv}

\usepackage{eccvabbrv}

\usepackage{graphicx}
\usepackage{booktabs}

\usepackage{bbm}
\usepackage{amsmath}
\usepackage{graphicx}
\usepackage{mathtools}
\usepackage{multirow}
\usepackage{colortbl}
\usepackage{adjustbox}
\usepackage{booktabs}
\usepackage[symbol]{footmisc}
\usepackage{array}
\newcolumntype{L}[1]{>{\raggedright\let\newline\\\arraybackslash\hspace{0pt}}m{#1}}
\newcolumntype{C}[1]{>{\centering\let\newline\\\arraybackslash\hspace{0pt}}m{#1}}
\newcolumntype{R}[1]{>{\raggedleft\let\newline\\\arraybackslash\hspace{0pt}}m{#1}}

\usepackage{wrapfig}
\usepackage{lipsum}
\usepackage{siunitx}
\usepackage{textcomp}
\usepackage{subcaption}
\usepackage{enumitem}
\usepackage{float}
\usepackage{geometry}

\newcommand{\red}[1]{\textcolor{red}{#1}}

\newcommand{\blue}[1]{\textcolor{blue}{#1}}

\usepackage[accsupp]{axessibility}  

\usepackage{hyperref}

\usepackage{orcidlink}

\begin{document}

\title{Rethinking the Role of Infrastructure \\
in Collaborative Perception}

\titlerunning{Rethinking the Role of Infrastructure in CP}

\author{Hyunchul Bae \inst{\dag}\and
Minhee Kang \inst{\dag}\and
Minwoo Song\and
Heejin Ahn \inst{*}}

\authorrunning{H.~Bae et al.}

\institute{Korea Advanced Institute of Science and Technology, Daejeon, South Korea \\ 
\email{\{bhc2675, ministop, haestle1, heejin.ahn\}@kaist.ac.kr}}

\maketitle
\renewcommand{\thefootnote}{\fnsymbol{footnote}}
\footnotetext[2]{These authors contributed equally to this work.}
\footnotetext[1]{Corresponding author.}

\begin{abstract}
Collaborative Perception (CP) is a process in which an ego agent receives and fuses sensor information from surrounding vehicles and infrastructure to enhance its perception capability. To evaluate the need for infrastructure equipped with sensors, extensive and quantitative analysis of the role of infrastructure data in CP is crucial, yet remains underexplored. To address this gap, we first quantitatively assess the importance of infrastructure data in existing vehicle-centric CP, where the ego agent is a vehicle. Furthermore, we compare vehicle-centric CP with infra-centric CP, where the ego agent is now the infrastructure, to evaluate the effectiveness of each approach. Our results demonstrate that incorporating infrastructure data improves 3D detection accuracy by up to 10.30\%, and infra-centric CP shows enhanced noise robustness and increases accuracy by up to 46.47\% compared with vehicle-centric CP.

\end{abstract}

\section{Introduction}\label{sec:intro}

The perception of autonomous vehicles has been enhanced with advancements in communication between vehicles and between vehicles and infrastructures. Because the communication enables vehicles and infrastructures to share their sensor information, an ego agent can fuse this shared data to broaden its perception areas and overcome occlusion issues \cite{xu2022v2xvit, hu2022where2comm}. This fusion technology is referred to as collaborative perception (CP).

Collaborative perception has recently garnered significant research attention \cite{xu2022cobevt, xu2022v2xvit, hu2022where2comm, wang2023core, yang2023scope, bae2024parcon}.  
Previous studies initially considered CP among vehicles \cite{xu2022cobevt, li2022v2x-sim} and recently started to include infrastructure data \cite{xu2022v2xvit, bae2024parcon}. Due to this historical reason, most previous studies use a vehicle as the ego agent and infrastructure as an auxiliary agent. Such a vehicle-centric approach may not take full advantage of the benefits of infrastructure, such as wider detection ranges and enhanced occlusion robustness \cite{fan2023calibration, yang2023bevheight, zimmer2023tumtraf}. This motivates us to evaluate the benefits of infrastructure in CP and rethink the role of infrastructure as the ego agent.

\setlength{\belowcaptionskip}{-15pt}
\begin{figure}[tb]
  \centering
  \begin{tabular}{cc}
       \includegraphics[width=0.48\textwidth]{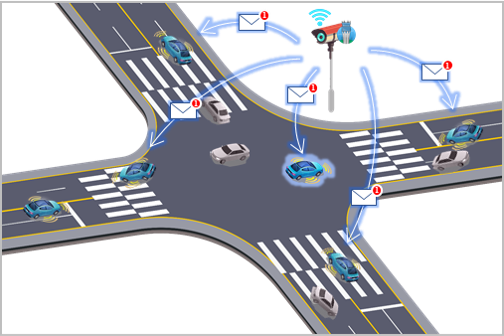} & 
       \includegraphics[width=0.48\textwidth]{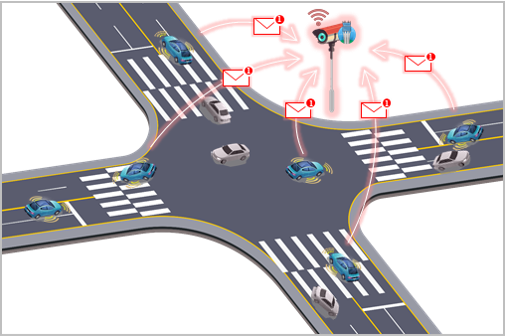} \\
       (a) Vehicle-centric CP & (b) Infra-centric CP
  \end{tabular}
  \caption{\textbf{About this paper.} We present quantitative analyses of the importance of infrastructure data in (a) vehicle-centric CP 
 and also in (b) infra-centric CP.}
  \label{fig1}
\end{figure}

In this study, we analyze the role of infrastructure in CP from two perspectives. First, we quantitatively analyze the importance of infrastructure data in existing vehicle-centric CP. Second, we compare vehicle-centric CP with \textit{infra-centric CP}, where infrastructure plays the role of the ego agent in CP. These two quantitative and extensive analyses are presented for the first time to the best of our knowledge.
Fig.~\ref{fig1} illustrates the concept of vehicle-centric CP and infra-centric CP. \\ 
\\ 

We summarize our contributions as follows.
\begin{enumerate}[label=(\roman*)]
    \item We perform quantitative analysis on the importance of infrastructure data in vehicle-centric CP in terms of 3D detection accuracy.
    \item We propose infra-centric CP and analyze specific scenarios where infra-centric CP is most effective.
    \item We extensively compare vehicle-centric CP and infra-centric CP in terms of 3D detection accuracy and noise sensitivity.
\end{enumerate}

The paper is structured as follows. In Sec.~\ref{sec:related}, we summarize previous studies of vehicle-centric and infra-centric CP. Sec.~\ref{sec:overview} overviews the process of CP, and Sec.~\ref{sec:experiments} presents the experiment details and results.
We conclude this paper in Sec.~\ref{sec:conclusion}.

\section{Related Works}\label{sec:related}

\subsubsection{Collaborative Perception.}
In CP, there are three main types of fusion methods: early fusion, late fusion, and intermediate fusion \cite{gao2024survey}. Early fusion refers to the fusion of raw sensor data and often requires high data bandwidth, making real-time computing difficult \cite{wulff2018early}. Late fusion refers to the fusion of the individual detection results, which doesn't require high computational power but yields the lowest perception performance \cite{yu2022dair}. To balance between the performance and computation, numerous studies utilize intermediate fusion as the primary strategy for CP \cite{xiang2023multi, xu2022v2xvit, hu2022where2comm}, where an ego agent fuses features, which are generated by feature extractor, to make prediction results. Since feature extraction and fusion often cause information loss or redundancy, suitable feature selection and fusion methods are crucial \cite{han2023collaborative}.

\subsubsection{Vehicle-Centric Collaborative Perception.}
Most previous studies in CP focus on vehicles as the ego agent \cite{xu2022cobevt, xu2022v2xvit, hu2022where2comm, wang2023core, yang2023scope}. In particular, vehicle-centric CP initially considers collaboration among multiple vehicles through vehicle-to-vehicle communication (V2V CP) \cite{li2022v2x-sim} and then includes infrastructure data through vehicle-to-everything communication (V2X CP) \cite{ngo2023cooperative}.
 By sharing the sensor data between vehicles, V2V CP significantly improves the perception performance through resolving occlusion issues \cite{wang2020v2vnet, xu2022opv2v}. 
 To utilize the advantages of infrastructure, such as having widened sensor range and robustness to the occlusion, V2X CP started to be considered \cite{lu2024extensible} \cite{yu2022dair}.
However, most existing studies have not yet extensively investigated the importance of infrastructure. In this paper, we quantitatively analyze the benefits of infrastructure data by comparing the performance between V2V CP and V2X CP.

\subsubsection{Infra-centric Collaborative Perception.}
For a standalone agent without collaboration, perception using infrastructure sensors shows higher performance compared to perception using vehicle sensors \cite{bai2022pillargrid}. Although there are various infrastructure-standalone perception studies, which emphasize the advantages of infrastructure sensors \cite{fan2023calibration, yang2023bevheight, zimmer2023tumtraf}, infra-centric CP remains underexplored. Even in previous studies in V2X CP, most have considered infrastructure as an auxiliary agent, not as the ego agent. Inspired by the need to explore infra-centric CP, we compare vehicle-centric and infra-centric CP, which uses infra-to-everything communication (I2X CP), by identifying suitable detection ranges for each approach and assessing accuracy and noise robustness. 

\section{Overview of Collaborative Perception Structure}\label{sec:overview}

\begin{figure}[bt]
  \centering
  \includegraphics[width=0.95\textwidth]{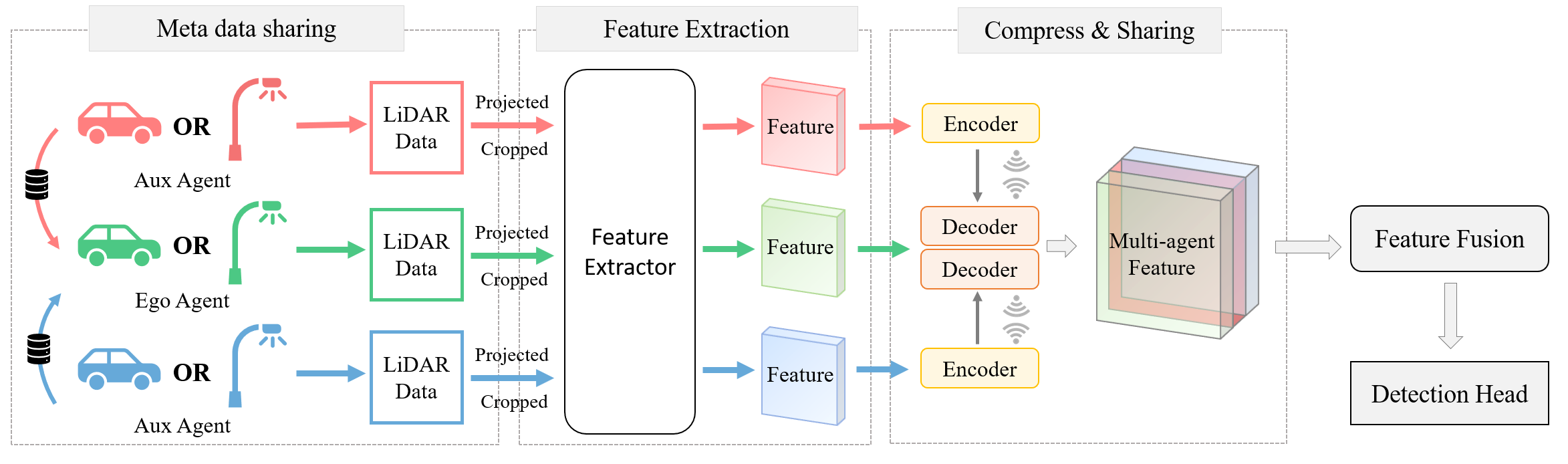}
   \caption{\textbf{Intermediate fusion of collaborative perception.} 
   We overview the most representative CP structure, consisting of metadata sharing, feature extraction, compress and sharing, feature fusion, and detection head.
   }
  \label{fig2}
\end{figure}

The overall architecture of CP, in particular intermediate fusion, is shown in Fig.~\ref{fig2}. The main processes are 1) Metadata Sharing, 2) Feature Extraction, 3) Feature Sharing, 4) Feature Fusion, and 5) Detection Head.

\setlength{\belowcaptionskip}{-15pt}

\subsubsection{Metadata Sharing.}
Agents in CP are divided into an \textit{ego agent} and \textit{auxiliary (aux) agents}. Aux agents are cooperative agents that provide complementary information to enable the ego agent to detect objects in broader regions. In the metadata sharing step, the ego agent receives the aux agents' metadata, including the type of agent, timestamp, and pose. There are two types of agents: vehicle ($\mathrm{V}$) and infrastructure ($\mathrm{I}$). 

\subsubsection{Feature Extraction.}
A feature is extracted from raw data through the encoder and the feature extractor. Different encoders are used for different types of raw data, such as images and pointclouds. In this paper, we focus only on the intermediate fusion of LiDAR and using PointPillars~\cite{lang2019pointpillars} and SECOND~\cite{yan2018second}, which are widely used encoders to refine pointclouds. Then, the feature is extracted from the encoded data through the feature extractor. The general feature extractor consists of the deep learning layers.

\subsubsection{Feature Sharing.}
The ego agent receives features shared by aux agents through communication. Feature sharing enables agents to exchange feature information. The features' size should be small enough to satisfy the limited transmission bandwidth and latency. Each aux agent compresses its feature, and the compressed features are provided to the ego agent. After the ego agent receives them, the ego agent decompresses the compressed features. 

\subsubsection{Feature Fusion.}
The ego agent fuses the shared feature information. 
Traditional fusion methods are concatenation \cite{liang2019mcfnet}, summation \cite{sun2019rtfnet}, and linear weighted \cite{arandjelovic2016weighted}. Recently, advanced fusion methods, such as graph-based fusion \cite{wang2020v2vnet} or attention-based fusion \cite{guo2022attention}, account for relationships and interactions among multiple agents. 

\subsubsection{Detection Head.}
The detection head takes the fused feature as the input and outputs the prediction of each object's center position $(x_c, y_c, z_c)$, size $(w, l, h)$, heading angle $\phi$, and label. These predicted values are used to calculate the loss. The loss is the sum of the regression loss and the classification loss. The regression loss is used to predict the position, size, and heading angle, and the classification loss is used to distinguish the object's label. In this paper, we use two labels: vehicle or not.

\section{Experiments}\label{sec:experiments}

In this section, we present the quantitative results of the importance of infrastructure data. In particular, we first explain datasets, existing vehicle-centric CP models, and implementation details. Then, we focus on the importance of infrastructure in vehicle-centric CP and infra-centric CP.

\subsection{Experiment Setup}

\subsubsection{Dataset.}
We utilize V2XSet \cite{xu2022v2xvit} and V2X-Sim \cite{li2021learning} datasets to validate the 3D object detection performance of vehicle-centric and infra-centric CP.  

V2XSet is a simulated dataset that supports V2X perception, co-simulated through CARLA \cite{dosovitskiy2017carla} and OpenCDA \cite{xu2021opencda}. It comprises 73 scenes featuring a minimum of 2 to 5 agents and incorporates 11,000 3D annotated LiDAR point cloud frames. The training, validation, and testing sets comprise 6.7K, 2K, and 2.8K frames, respectively. The training set contains 33 scenarios (18 V2V scenarios, 15 V2X scenarios), and the validation and testing sets contain 25 scenarios (11 V2V scenarios, 14 V2X scenarios). We refer to the entire V2XSet dataset as V2XSet-W and the selected dataset to contain only V2X scenarios as V2XSet-I.

V2X-Sim is an extensive simulated dataset focused on multi-agent perception. It is generated using the traffic simulation software SUMO \cite{krajzewicz2012recent} and CARLA \cite{dosovitskiy2017carla}. V2X-Sim collects 100 scenes, each consisting of 10,000 frames, utilizing RGB cameras, LiDAR, GPS, and IMU, with 2-5 vehicles and 1 infrastructure in each scene. The training, validation, and testing sets comprise 8K, 1K, and 1K frames, respectively.

\subsubsection{Models.}
We select models based on the following two factors: \lowercase\expandafter{\romannumeral1}) Models should be published within the past two years.  \lowercase\expandafter{\romannumeral2}) Models should be widely referenced in related works. \lowercase\expandafter{\romannumeral3}) Models can be changed to infra-centric CP without significant structure modification.
Considering these factors, we select V2X-ViT~\cite{xu2022v2xvit}, Where2comm~\cite{hu2022where2comm}, and ParCon~\cite{bae2024parcon}. 
\subsubsection{Implementation Details.} \label{sec:implement_details}
We train all models using AdamW~\cite{loshchilov2017adamw}, with a learning rate of 3e-4 and a weight decay of 0.01. Also, we use Cosine Annealing Warm-Up Restarts~\cite{loshchilov2016sgdr}, with a warm-up learning rate of 2e-4 and warm-up epochs of 10. Each model trains up to 40 epochs on an RTX 4090. We train the model with three different settings: perfect, simple noise, and harsh noise. The details of these settings are given in \cite{bae2024parcon}.

We modify the CP models to convert vehicle-centric CP to infra-centric CP. For all the dataset scenarios, we set the infrastructure as the ego agent and change the the detection range of $z$ to consider the height difference between the vehicle and the infrastructure. In V2XSet-I, we convert the range of $z$ from $z\in[-3, 1]$ to $z\in[-5, -1]$, and, in V2X-Sim, we convert from $z\in[-3, 2]$ to $z\in[-8.5, -3.5]$. We adjust the detection range of $x$ and $y$ of the ego agent. In vehicle-centric CP,
we use the default detection range given in \cite{xu2022v2xvit} for V2XSet and \cite{li2022v2x-sim} for V2X-Sim. In detail, the detection range of V2XSet-W and V2XSet-I is $x\in[-140.8, 140.8]$ and $y\in[-38.4, 38.4]$, and the detection range of V2X-Sim is $x\in[-32, 32]$ and $y\in[-32, 32]$. In infra-centric CP, we only change the detection range in V2XSet-I. We use the square-shaped detection range of $x\in[-76.8, 76.8]$ and $y\in[-76.8, 76.8]$. The reason for this selection is explained in Sec.~\ref{exp:detection_range}. In V2X-Sim, we set the detection range for infra-centric CP as the same as vehicle-centric CP because the default detection range is already square-shaped.

In summary, the detection range is different depending on the dataset. We interpret that V2XSet makes CP models detect broadened regions, e.g., covering more than two intersections, and V2X-Sim makes CP models detect nearby regions, e.g., covering one intersection.

\subsection{Study of Infrastructure Data in Vehicle-centric CP}
In this section, we aim to demonstrate the usefulness of infrastructure data in vehicle-centric CP. To do this, we compare two vehicle-centric CPs, V2V CP and V2X CP.

\subsubsection{Dataset Details.} \label{sec:v2v_v2x_dataset}
We train vehicle-centric CP models of V2X-ViT, Where2comm, and ParCon with the V2XSet-W dataset. 
In the validation and test sets, we have chosen 12 scenarios involving two vehicles and one infrastructure. We use the same vehicle as the ego agent but an aux vehicle agent for V2V CP and an aux infrastructure agent for V2X CP. 
We also apply the same approach to the experiment with the V2X-Sim dataset. We use 10 scenarios for validation. We set the maximum number of agents as 4 and randomly choose vehicles in every scenario. Then, the chosen vehicles are used for V2V CP, and we swap one aux vehicle to an aux infrastructure for V2X CP.

\begin{table}[t!]
\centering
\scriptsize
\setlength{\tabcolsep}{1pt}
\caption{Comparison of AP@0.7 accuracy using infrastructure data. V2V means the types of an ego agent and aux agents are vehicles, and V2X means the type of an ego agent is a vehicle, and the types of aux agents are vehicles or infrastructure.}
\label{tab:accuracy_v2v_v2x}
\renewcommand{\arraystretch}{1.1}
\begin{tabular}{l|C{1cm}C{1cm}C{1cm}C{1cm}|C{1cm}C{1cm}C{1cm}C{1cm}}
\hline
\rowcolor[HTML]{D7D5D5} 
\multicolumn{1}{c|}{\cellcolor[HTML]{D7D5D5}}& \multicolumn{4}{c|}{\cellcolor[HTML]{D7D5D5}V2XSet}& \multicolumn{4}{c}{\cellcolor[HTML]{D7D5D5}V2X-Sim}\\
\rowcolor[HTML]{D7D5D5} 
\multicolumn{1}{c|}{\cellcolor[HTML]{D7D5D5}}& \multicolumn{2}{c|}{\cellcolor[HTML]{D7D5D5}Perfect}& \multicolumn{2}{c|}{\cellcolor[HTML]{D7D5D5}Simple Noise} & \multicolumn{2}{c|}{\cellcolor[HTML]{D7D5D5}Perfect}& \multicolumn{2}{c}{\cellcolor[HTML]{D7D5D5}Simple Noise} \\
\rowcolor[HTML]{D7D5D5} 
\multicolumn{1}{c|}{\multirow{-3}{*}{\cellcolor[HTML]{D7D5D5}Model}} & \multicolumn{1}{C{1cm}|}{\cellcolor[HTML]{D7D5D5}V2V} & \multicolumn{1}{C{1cm}|}{\cellcolor[HTML]{D7D5D5}V2X} & \multicolumn{1}{C{1cm}|}{\cellcolor[HTML]{D7D5D5}V2V}   & V2X  & \multicolumn{1}{C{1cm}|}{\cellcolor[HTML]{D7D5D5}V2V} & \multicolumn{1}{C{1cm}|}{\cellcolor[HTML]{D7D5D5}V2X} & \multicolumn{1}{C{1cm}|}{\cellcolor[HTML]{D7D5D5}V2V}  & V2X  \\ \hline
No Fusion   & \multicolumn{1}{c|}{0.540}& \multicolumn{1}{c|}{0.540}& \multicolumn{1}{c|}{0.540}&0.540& \multicolumn{1}{c|}{0.517}& \multicolumn{1}{c|}{0.517}& \multicolumn{1}{c|}{0.517}&0.517\\ \hline
V2X-ViT \cite{xu2022v2xvit}     & \multicolumn{1}{c|}{0.673}& \multicolumn{1}{c|}{0.722}& \multicolumn{1}{c|}{0.548}&0.564& \multicolumn{1}{c|}{0.745}& \multicolumn{1}{c|}{0.775}& \multicolumn{1}{c|}{0.578}&0.594\\
Where2comm \cite{hu2022where2comm}  & \multicolumn{1}{c|}{0.713}& \multicolumn{1}{c|}{0.785}& \multicolumn{1}{c|}{0.572}&0.580& \multicolumn{1}{c|}{0.733}& \multicolumn{1}{c|}{0.752}& \multicolumn{1}{c|}{0.559}&0.555\\
ParCon \cite{bae2024parcon}       & \multicolumn{1}{c|}{0.734}& \multicolumn{1}{c|}{0.810}& \multicolumn{1}{c|}{0.630}&0.646& \multicolumn{1}{c|}{0.793}& \multicolumn{1}{c|}{0.829}& \multicolumn{1}{c|}{0.615}&0.621\\ \hline
\end{tabular}
\vspace{-10pt}
\end{table}

\subsubsection{Accuracy.} \label{sec:v2v_v2x_accuracy}


The results of the accuracy comparison between V2V CP and V2X CP are in Tab.~\ref{tab:accuracy_v2v_v2x}. Regarding the detection accuracy on V2XSet, for all the models, V2X CP shows better accuracy than V2V CP. In the perfect setting, AP@0.7 increases by between 7.29\% and 10.30\%, and in the simple noise setting, AP@0.7 increases by between 1.43\% and 2.98\%. Regarding the detection accuracy on V2X-Sim, for all the models, V2X CP shows better accuracy in the perfect setting, enhanced by between 2.53\% and 4.55\%, and in the simple noise setting, AP@0.7 of V2X-ViT and ParCon increase by 2.71\% and 0.86\%, respectively. However, Where2comm has lower accuracy about V2X CP than V2V CP, decreased by 0.73\%.

In the simple noise setting, the accuracy of V2X CP is better in V2XSet and worse in V2X-Sim than that of V2V CP. The main cause is noise itself. Noise tends to be more severe when sensor data points are far from the sensor (e.g., for the same level of heading angle noise, data points farther from the sensor represent excessive movement compared to those closer to the sensor). For this reason, the infrastructure's property of a broader sensor range makes the infrastructure more vulnerable to noise than the vehicle. Furthermore, noise makes perception ability more vulnerable with the smaller detection range. As mentioned in Sec.~\ref{sec:implement_details}, V2X-Sim uses a smaller detection range than V2XSet. The smaller detection range is less likely to include the auxiliary agents' data. This indicates that the infrastructure might rarely influence the ego vehicle to enhance perception, or infrastructure data with noise deteriorates the perception ability, which acts as an obstruction.

\begin{table}[b!]
\centering
\vspace{-10pt}
\scriptsize
\setlength{\tabcolsep}{1pt}
\caption{Comparison of AP@0.7 accuracy using infrastructure data across different scenarios. Difference means the change from AP@0.7 in V2V to AP@0.7 in V2X.}
\label{tab:scenario_accuracy_v2v_v2x}
\vspace{-7pt}
\renewcommand{\arraystretch}{1.1}
\begin{tabular}{l|C{1.3cm}|C{1.3cm}|C{1.3cm}|C{1.3cm}|C{1.3cm}|C{1.3cm}|C{1.3cm}}
\hline
\rowcolor[HTML]{D7D5D5} 
\multicolumn{1}{c|}{\cellcolor[HTML]{D7D5D5}Model} & CP       & Scene \#1      & Scene \#3      & Scene \#4      & Scene \#8      & Scene \#9      & Scene \#11 \\ \hline
\multirow{3}{*}{V2X-ViT\cite{xu2022v2xvit}}    &  V2V          &0.925&0.852&0.579&0.634&0.668&0.768\\
& V2X          &0.907&0.742&0.776&0.698&0.724&0.746\\
                            &Difference&\red{0.018$\downarrow$}&\red{0.110$\downarrow$}&\blue{0.197$\uparrow$}&\blue{0.064$\uparrow$}&\blue{0.056$\uparrow$}&\red{0.022$\downarrow$}\\  \hline
\multirow{3}{*}{Where2comm\cite{hu2022where2comm}}  & V2V          &0.954&0.889&0.602&0.648&0.690&0.818\\
                            & V2X          &0.956&0.804&0.809&0.733&0.786&0.822\\
                            &Difference&\blue{0.002$\uparrow$}&\red{0.085$\downarrow$}&\blue{0.208$\uparrow$}&\blue{0.084$\uparrow$}&\blue{0.097$\uparrow$}&\blue{0.005$\uparrow$}\\
                            \hline
\multirow{3}{*}{ParCon\cite{bae2024parcon}}      & V2V&0.953&0.892&0.680&0.710&0.699&0.865\\
                            & V2X&0.971&0.857&0.854&0.798&0.818&0.854\\
                            &Difference&\blue{0.019$\uparrow$}&\red{0.035$\downarrow$}&\blue{0.174$\uparrow$}&\blue{0.089$\uparrow$}&\blue{0.120$\uparrow$}&\red{0.012$\downarrow$}\\ \hline
\end{tabular}
\end{table}

\begin{figure}[t!]
\vspace{5pt}
  \centering
  \begin{tabular}{cc}
       \includegraphics[width=0.47\textwidth]{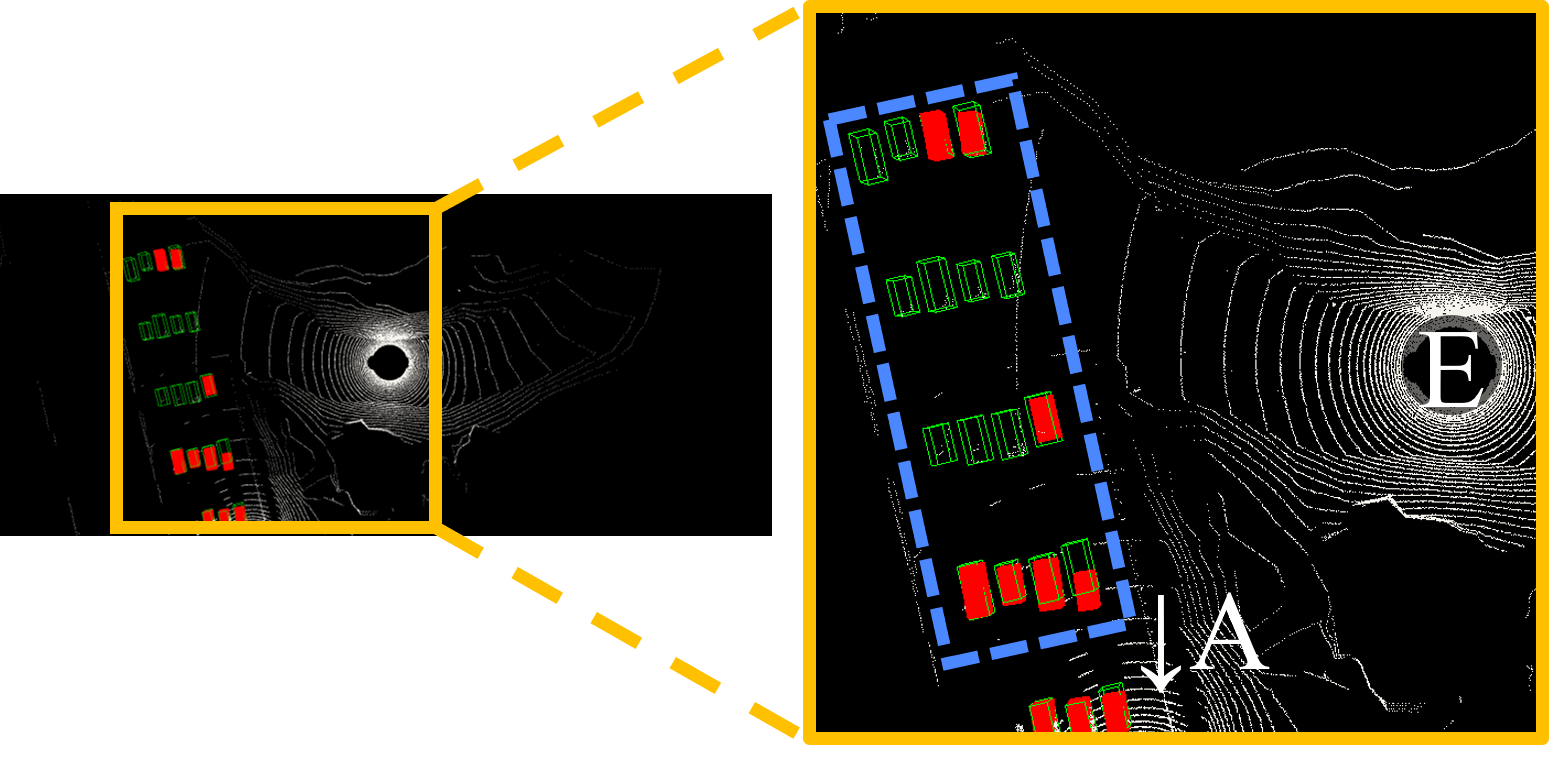} & 
       \includegraphics[width=0.47\textwidth]{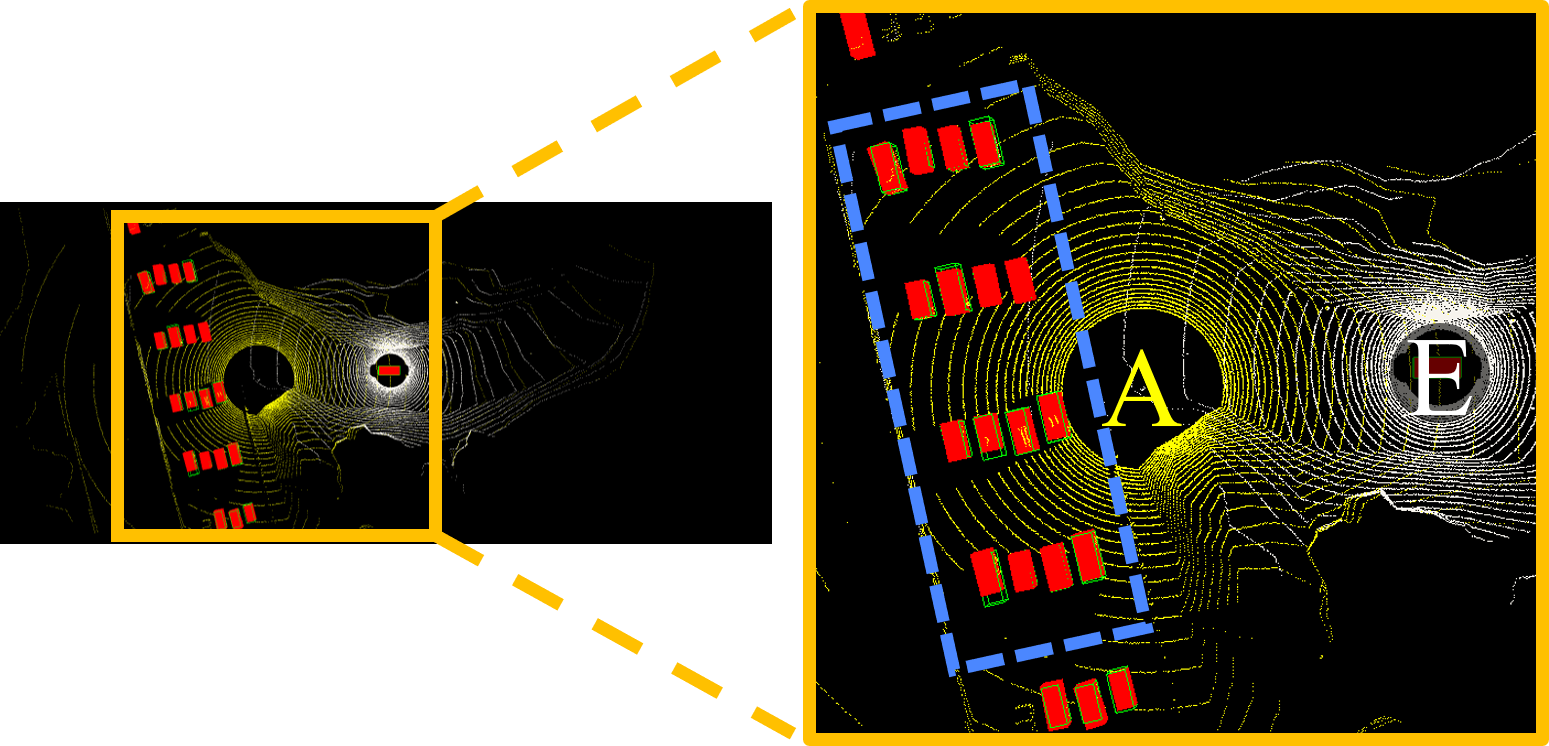} \\
       (a) V2V CP & (b) V2X CP
   \end{tabular}
   \vspace{-7pt}
   \caption{\textbf{Effective case of infrastructure data (Scene \#4).} The point clouds from the vehicle's LiDAR are indicated as white dots, and the point clouds from the infrastructure's LiDAR are indicated as yellow dots. The green bounding boxes are ground truth objects within the ego agent's detection range. \textbf{E} means the ego agent and \textbf{A} means an aux agent.}
   \vspace{5pt}
   \label{fig:Scenario_Infra_Data_Best}
\end{figure}

\begin{figure}[t!]
  \centering
  \begin{tabular}{cc}
       \includegraphics[width=0.47\textwidth]{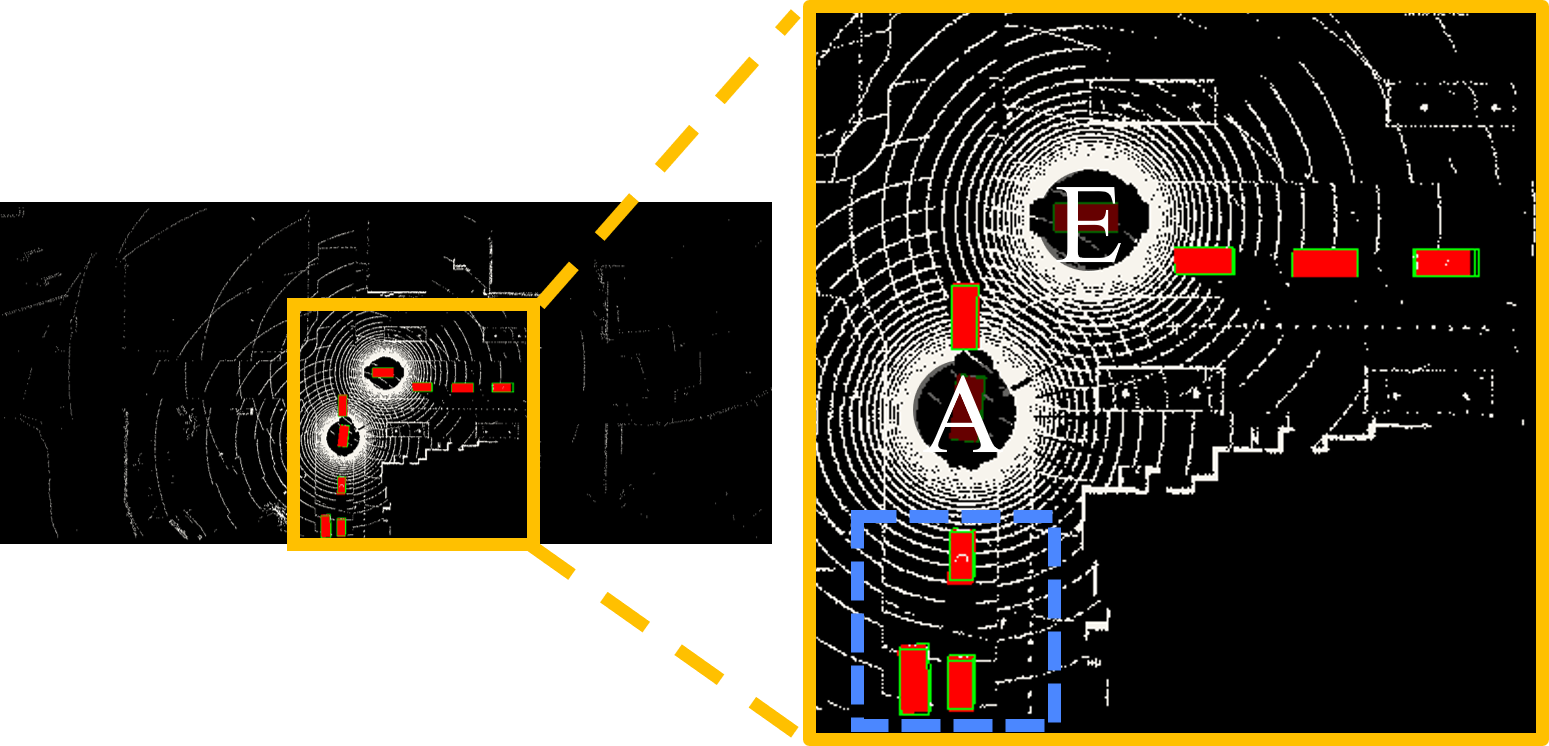} & 
       \includegraphics[width=0.47\textwidth]{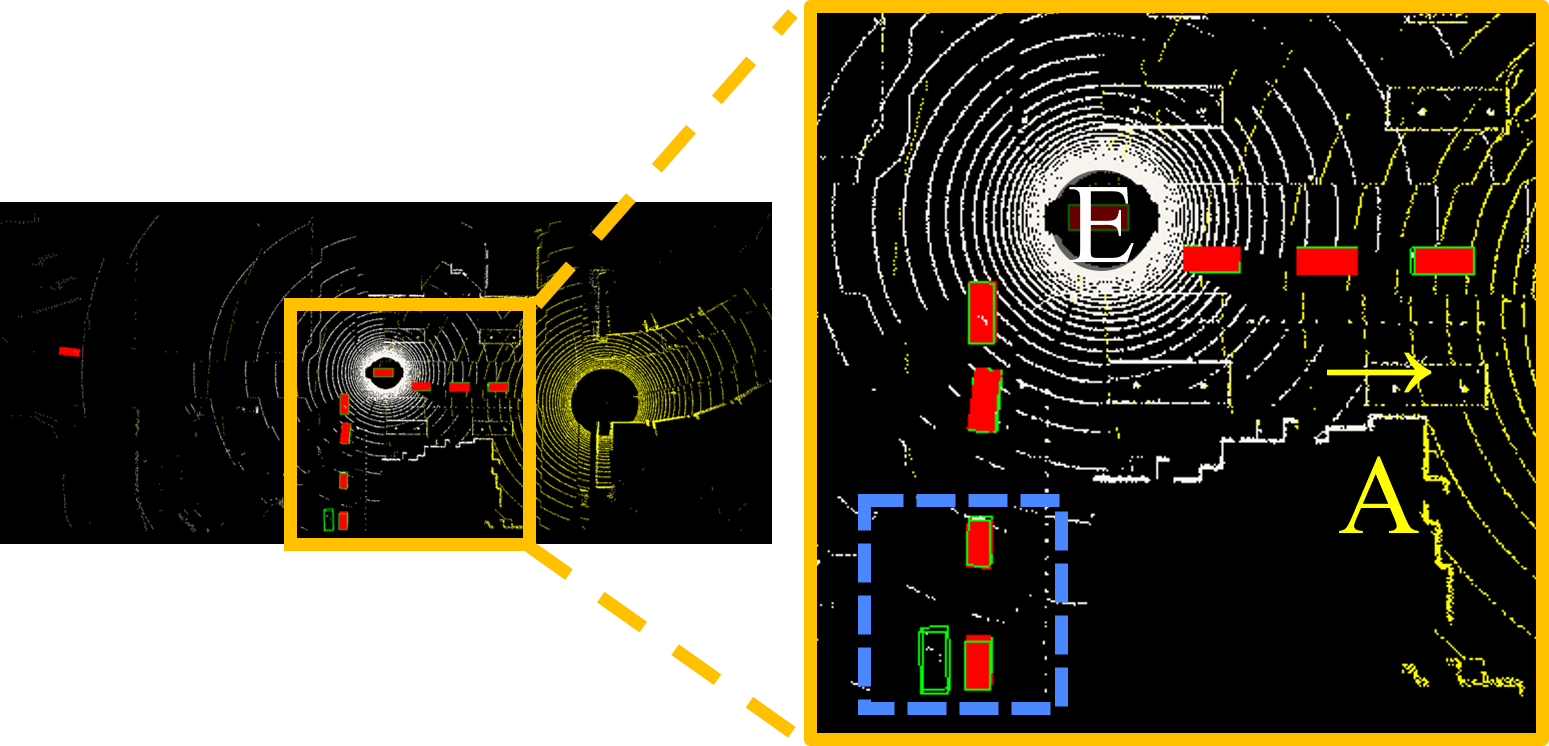} \\
       (a) V2V Collaborative Perception & (b) V2X Collaborative Perception 
  \end{tabular}
   \vspace{-7pt}
   \caption{\textbf{Uneffective case of infrastructure data (Scene \#3).} The point clouds from the vehicle's LiDAR are indicated as white dots, and the point clouds from the infrastructure's LiDAR are indicated as yellow dots. The green bounding boxes are ground truth objects within the ego agent's detection range. \textbf{E} means the ego agent and \textbf{A} means an aux agent.}
  \label{fig:Scenario_Infra_Data_Worst}
\end{figure}

\subsubsection{Scenarios Analysis.}
We compare V2V CP and V2X CP in the same specific scenarios on the V2XSet dataset. As shown in Tab.~\ref{tab:scenario_accuracy_v2v_v2x}, the performance of V2V CP and V2X CP largely depends on scenarios. To analyze when V2X CP is more effective than V2V CP, we select two scenarios, Scene \#4 and Scene \#3, which are the most and least effective cases of V2X CP, respectively. 

The effects of infrastructure data are significant at 4-way and 3-way urban intersections and merging sections in a freeway. As shown in a merging section scenario in Fig.~\ref{fig:Scenario_Infra_Data_Best}, in V2V CP, the aux vehicle does not contribute to providing information on objects on the main road due to occlusions. In contrast, in V2X CP, the aux infrastructure can detect all objects on the main road. This result demonstrates the importance of infrastructure data when occlusions occur due to heavy traffic.

On the other hand, the effects of infrastructure data may degrade the performance in certain scenarios, such as two intersection environments, as shown in Fig.~\ref{fig:Scenario_Infra_Data_Worst}. In this environment, infrastructure only observes objects in one intersection and cannot provide information on objects in the other intersection. When the detection range of the ego agent is far outside of the sensor range of infrastructure, infrastructure data may not be useful.


\subsection{Study of Infra-centric CP}
This section aims to suggest infra-centric CP, which sets infrastructure as the ego agent. We also use the term {I2X CP} to indicate infra-centric CP to differentiate it from V2X CP. We first study the effect of detection ranges of the ego agent and compare detection accuracy and noise sensitivity.
We perform three experiments: 1) Finding a suitable detection range with V2XSet-I, 2) The results of accuracy and noise sensitivity with V2XSet-I and V2X-Sim, and 3) The analysis of scenarios on V2XSet-I.

\begin{table}[b!]
\centering
\scriptsize
\begin{tabular}{cc}
        \parbox[c][3cm][c]{0.43\textwidth}{
        \centering
        \scriptsize
        \setlength{\tabcolsep}{1.5pt}
        \renewcommand{\arraystretch}{1.1}
        \caption{Effect of detection range in V2X CP on the V2XSet-I dataset.}
        \label{tab:detection_range_V2X}
        \vspace{-7pt}
        \begin{tabular}{l|c|c}
        \hline
        \multicolumn{1}{c|}{\cellcolor[HTML]{D7D5D5}}                        & \cellcolor[HTML]{D7D5D5}Rectangle  & \cellcolor[HTML]{D7D5D5}Square     \\
        \multicolumn{1}{c|}{\multirow{-2}{*}{\cellcolor[HTML]{D7D5D5}Model}} & \cellcolor[HTML]{D7D5D5}AP@0.5/0.7 & \cellcolor[HTML]{D7D5D5}AP@0.5/0.7 \\ \hline
        V2X-ViT \cite{xu2022v2xvit}&\textbf{0.926/0.875}&0.899/0.815\\
        Where2comm \cite{hu2022where2comm}&\textbf{0.945/0.913}&0.933/0.881\\
        ParCon\cite{bae2024parcon}&0.938/0.902&\textbf{0.944/0.904}\\\hline
        \end{tabular}
        } & \quad \quad
        \parbox[c][3cm][c]{0.43\textwidth}{
        \centering
        \scriptsize
        \setlength{\tabcolsep}{1.5pt}
        \renewcommand{\arraystretch}{1.1}
        \caption{Effect of detection range in I2X CP on the V2XSet-I dataset.}
        \label{tab:detection_range_I2X}
        \vspace{-7pt}
        \begin{tabular}{l|c|c}
        \hline
        \multicolumn{1}{c|}{\cellcolor[HTML]{D7D5D5}}                        & \cellcolor[HTML]{D7D5D5}Rectangle  & \cellcolor[HTML]{D7D5D5}Square     \\
        \multicolumn{1}{c|}{\multirow{-2}{*}{\cellcolor[HTML]{D7D5D5}Model}} & \cellcolor[HTML]{D7D5D5}AP@0.5/0.7 & \cellcolor[HTML]{D7D5D5}AP@0.5/0.7 \\ \hline
        V2X-ViT \cite{xu2022v2xvit}&0.890/0.828&\textbf{0.915/0.882}\\
        Where2comm \cite{hu2022where2comm}&0.882/0.800&\textbf{0.947/0.910}\\
        ParCon\cite{bae2024parcon}&0.885/0.818&\textbf{0.936/0.908}\\ \hline
        \end{tabular}
        }
    \end{tabular}
\end{table}

\begin{figure}[b!]
  \centering
  \begin{tabular}{c}
       \includegraphics[width=0.98\textwidth]{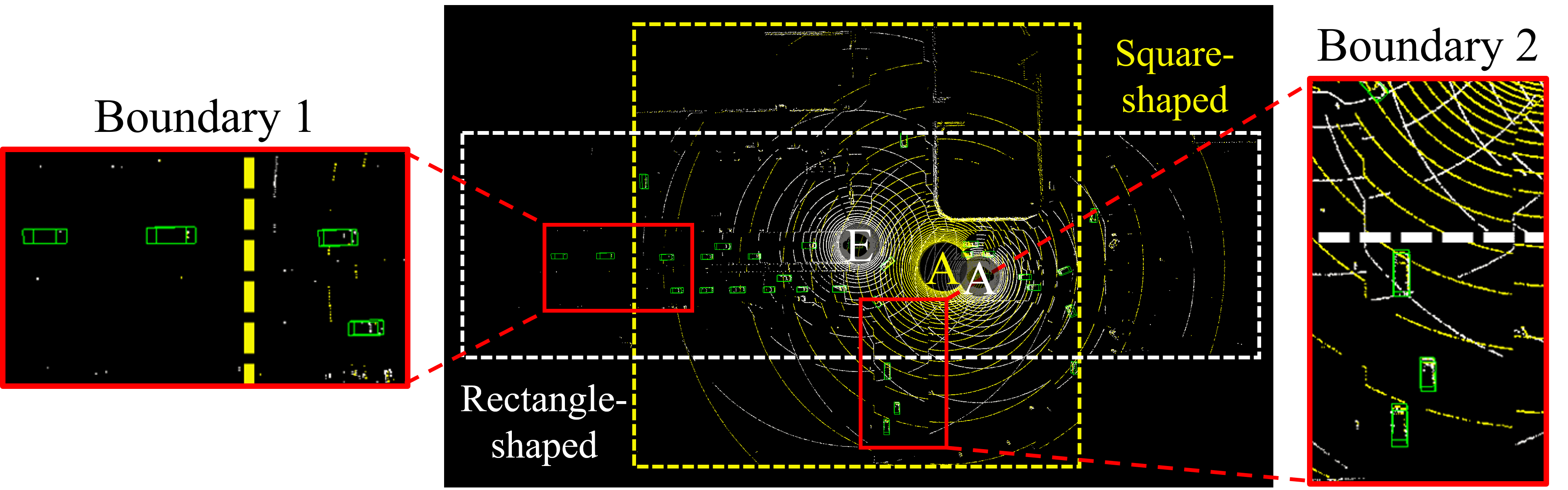} \\
       (a) V2X Collaborative Perception\\
       \\
       
       \includegraphics[width=0.98\textwidth]{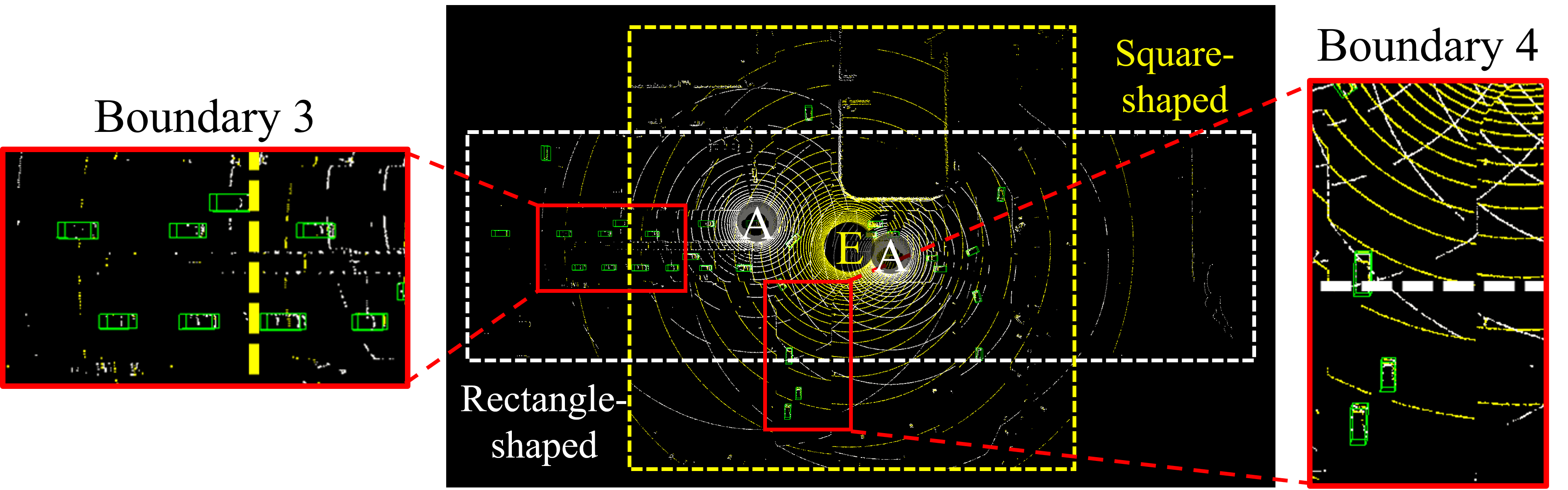} \\
       (b) I2X Collaborative Perception 
  \end{tabular}
   \caption{\textbf{Comparison between Shapes of the Detection Range of the Ego Agent.} The point clouds from the vehicle's LiDAR are indicated as white dots, and the point clouds from the infrastructure's LiDAR are indicated as yellow dots. The green bounding boxes are ground truth objects in the ego agent's detection range. \textbf{E} means the ego agent and \textbf{A} means aux agents. The white dotted line indicates the rectangle-shaped detection range, and the yellow dotted line indicates the square-shaped detection range.}
  \label{fig:detection_range}
\end{figure}

\subsubsection{Study of Detection Range.} \label{exp:detection_range}

Depending on the type of the ego agent, the detection range may have to change. As shown in Tab.~\ref{tab:detection_range_V2X} and Tab.~\ref{tab:detection_range_I2X}, the rectangle shape of the detection range shows better performance than the square shape in V2X CP. On the other hand, the square shape performs better than the rectangle shape in I2X CP. In Fig.~\ref{fig:detection_range} (a), the detection range for the ego vehicle in V2X CP should adopt a rectangular shape, aligning with the road's layout and the vehicle's forward movement. When the square-shaped detection range is applied, the heading direction length is shorter than the rectangle-shaped detection range, thereby missing the objects far away from the ego vehicle along the heading direction (\textbf{Boundary 1} in Fig.~\ref{fig:detection_range} (a)). Also, the square-shaped detection range makes V2X CP detect unnecessary objects that are unlikely to interact with the ego vehicle (\textbf{Boundary 2} in Fig.~\ref{fig:detection_range} (a)).
Conversely, in Fig.~\ref{fig:detection_range} (b), the detection range of the ego infrastructure in I2X CP should be square-shaped. This range leverages the property that infrastructure sensors are elevated and can monitor all directions equally. When the rectangle-shaped detection range is applied, it includes an area that is not of interest (\textbf{Boundary 3} in Fig.~\ref{fig:detection_range} (b)). Also, the rectangle-shaped detection range makes I2X CP unable to fully observe the responsible intersection area (\textbf{Boundary 4} in Fig.~\ref{fig:detection_range} (b)). 

\subsubsection{Dataset Details.} \label{sec:dataset_details_v2x_i2x}
Regarding V2XSet, the models in this section are trained using V2XSet-I, and the type of ego agent changes depending on the type of CP (V2X or I2X). Based on the study of the detection range, V2X CP uses the rectangle-shaped detection range of $x\in[-140.8, 140.8]$ and $y\in[-38.4, 38.4]$, and I2X CP uses the square-shaped detection range of $x\in[-76.8, 76.8]$ and $y\in[-76.8, 76.8]$. Regarding V2X-Sim, the models are trained and validated with the original training and validation datasets, respectively. We use the default square-shaped detection range, $x\in[-32, 32]$ and $y\in[-32, 32]$, for both V2X CP and I2X CP. V2X CP uses an ego vehicle and aux vehicles/infrastructure, and I2X CP uses an ego infrastructure and aux vehicles.

\subsubsection{Accuracy.}

To identify the effects of the type of an ego agent, we compare the detection accuracy of V2X CP and I2X CP as shown in Tab.~\ref{tab:accuracy_v2x_i2x}. In the perfect setting on the V2XSet, all the models show that I2X CP outperforms V2X CP in accuracy. Also, I2X CP is more effective in the simple noise setting, showing a rate of increase from 16.65\% to 19.13\% on V2XSet and from 37.51\% to 46.47\% on V2X-Sim. Unlike the comparison between V2V and V2X CP in sec.~\ref{sec:v2v_v2x_accuracy}, it is noteworthy that I2X CP always exhibits better performance in the simple noise setting at AP@0.7. The finding indicates that the ego infrastructure has noise robustness. The standalone infrastructure performs well, and based on this infrastructure's ability, I2X CP maintains its performance even in the noise. Regarding the comparison between datasets, the change rate in V2X-Sim is significantly larger than that in V2XSet. As we mentioned in Sec.~\ref{sec:implement_details}, the phenomenon might stem from the detection range of V2X-Sim, which is almost square-shaped and corresponds with a region at an intersection. Thus, I2X CP can cover almost every region within the detection range by minimizing occlusion. This supports our logic in shaping the detection range in Sec.~\ref{exp:detection_range}.

\begin{table}[t!]
\centering
\scriptsize
\setlength{\tabcolsep}{1pt}
\caption{Comparison of AP@0.7 accuracy based on ego agent type. V2X means the type of an ego agent is a vehicle, and I2X means an ego agent is an infrastructure.}
\label{tab:accuracy_v2x_i2x}
\vspace{-7pt}
\renewcommand{\arraystretch}{1.1}
\begin{tabular}{l|C{1cm}C{1cm}C{1cm}C{1cm}|C{1cm}C{1cm}C{1cm}C{1cm}}
\hline
\rowcolor[HTML]{D7D5D5} 
\multicolumn{1}{c|}{\cellcolor[HTML]{D7D5D5}}& \multicolumn{4}{c|}{\cellcolor[HTML]{D7D5D5}V2XSet}& \multicolumn{4}{c}{\cellcolor[HTML]{D7D5D5}V2X-Sim}\\
\rowcolor[HTML]{D7D5D5} 
\multicolumn{1}{c|}{\cellcolor[HTML]{D7D5D5}}& \multicolumn{2}{c|}{\cellcolor[HTML]{D7D5D5}Perfect}& \multicolumn{2}{c|}{\cellcolor[HTML]{D7D5D5}Simple Noise} & \multicolumn{2}{c|}{\cellcolor[HTML]{D7D5D5}Perfect}& \multicolumn{2}{c}{\cellcolor[HTML]{D7D5D5}Simple Noise} \\
\rowcolor[HTML]{D7D5D5} 
\multicolumn{1}{c|}{\multirow{-3}{*}{\cellcolor[HTML]{D7D5D5}Model}} & \multicolumn{1}{C{1cm}|}{\cellcolor[HTML]{D7D5D5}V2X} & \multicolumn{1}{C{1cm}|}{\cellcolor[HTML]{D7D5D5}I2X} & \multicolumn{1}{C{1cm}|}{\cellcolor[HTML]{D7D5D5}V2X}   & I2X  & \multicolumn{1}{C{1cm}|}{\cellcolor[HTML]{D7D5D5}V2X} & \multicolumn{1}{C{1cm}|}{\cellcolor[HTML]{D7D5D5}I2X} & \multicolumn{1}{C{1cm}|}{\cellcolor[HTML]{D7D5D5}V2X}  & I2X  \\ \hline
No Fusion   & \multicolumn{1}{c|}{0.602}& \multicolumn{1}{c|}{\textbf{0.718}}& \multicolumn{1}{c|}{0.602}&\textbf{0.718}& \multicolumn{1}{c|}{0.517}& \multicolumn{1}{c|}{\textbf{0.795}}& \multicolumn{1}{c|}{0.517}&\textbf{0.795}\\ \hline
V2X-ViT \cite{xu2022v2xvit}    & \multicolumn{1}{c|}{0.875}& \multicolumn{1}{c|}{\textbf{0.882}}& \multicolumn{1}{c|}{0.716}&\textbf{0.853}& \multicolumn{1}{c|}{0.775}& \multicolumn{1}{c|}{\textbf{0.879}}& \multicolumn{1}{c|}{0.594}&\textbf{0.820}\\
Where2comm \cite{hu2022where2comm} & \multicolumn{1}{c|}{\textbf{0.913}}& \multicolumn{1}{c|}{0.910}& \multicolumn{1}{c|}{0.724}&\textbf{0.855}& \multicolumn{1}{c|}{0.752}& \multicolumn{1}{c|}{\textbf{0.868}}& \multicolumn{1}{c|}{0.555}&\textbf{0.813}\\
ParCon \cite{bae2024parcon}      & \multicolumn{1}{c|}{0.902}& \multicolumn{1}{c|}{\textbf{0.908}}& \multicolumn{1}{c|}{0.737}&\textbf{0.859}& \multicolumn{1}{c|}{0.829}& \multicolumn{1}{c|}{\textbf{0.896}}& \multicolumn{1}{c|}{0.621}&\textbf{0.853}\\ \hline
\end{tabular}
\vspace{-16pt}
\end{table}

\begin{figure}[t!]
\vspace{-5pt}
\includegraphics[width=1.0\textwidth]{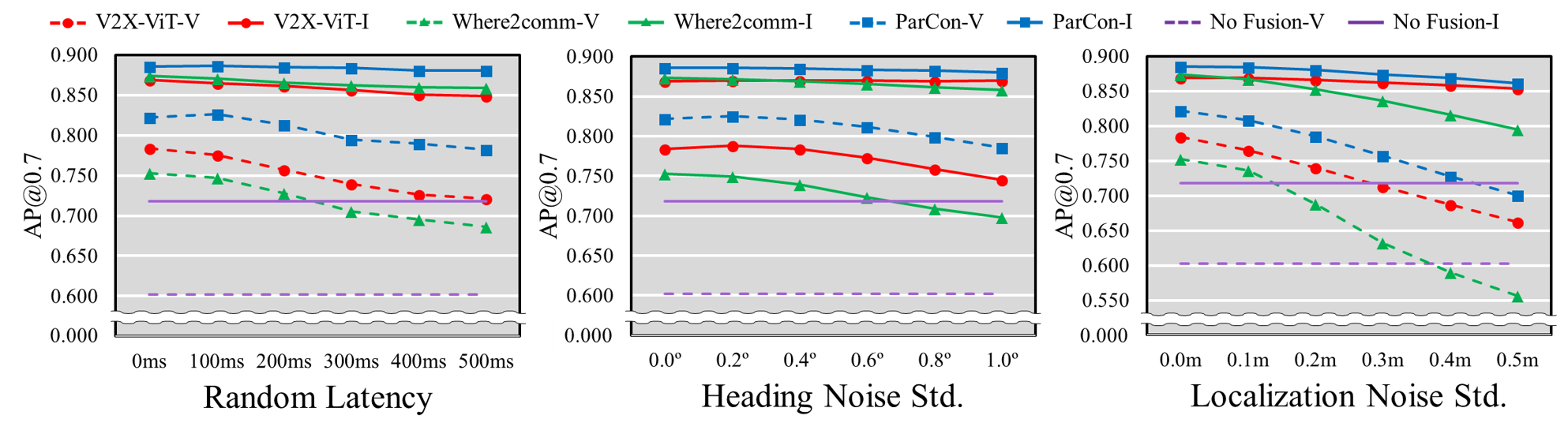}
\vspace{-20pt}
   \caption{\textbf{Noise Sensitivity Comparison on V2XSet.} Vehicle-centric CP is referred to as ``-V'' and infra-centric as ``-I.'' Vehicle-standalone perception is referred to as ``No Fusion-V'' and infrastructure-standalone perception as ``No Fusion-I.''}
  \label{fig:harsh_noise_V2XSet}
\end{figure}

\begin{figure}[t!]
\vspace{10pt}
\includegraphics[width=1.0\textwidth]{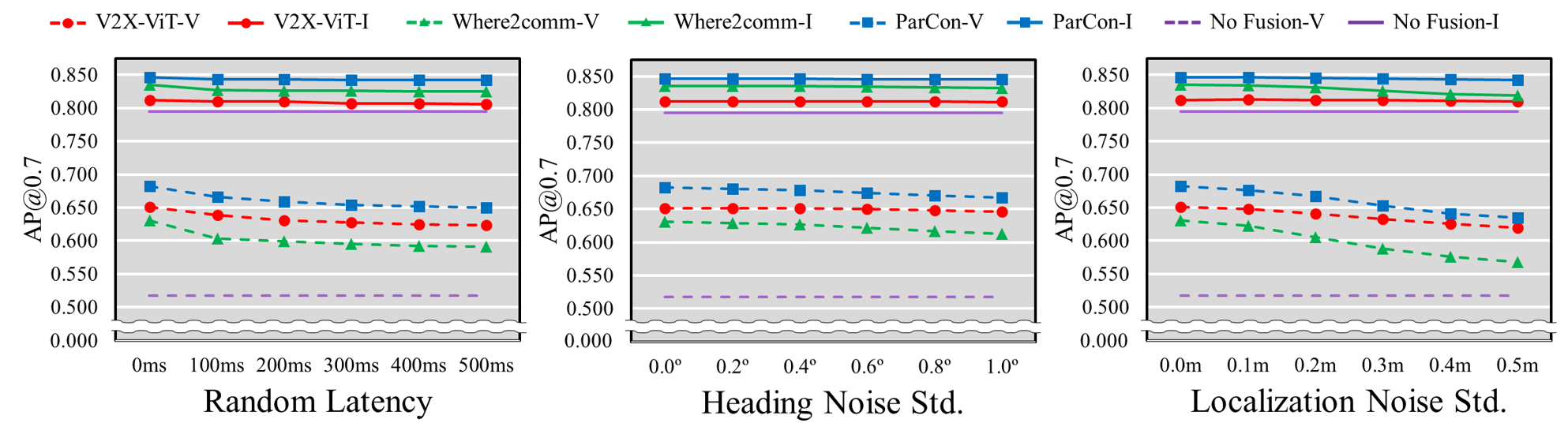}
\vspace{-20pt}
\caption{\textbf{Noise Sensitivity Comparison on V2X-Sim.} Vehicle-centric CP is referred to as ``-V'' and infra-centric as ``-I.'' Vehicle-standalone perception is referred to as ``No Fusion-V'' and infrastructure-standalone perception as ``No Fusion-I.''}
  \label{fig:harsh_noise_V2X-Sim}
\end{figure}

\subsubsection{Noise Sensitivity.}
Based on the model trained with the harsh noise setting, we compare the noise sensitivity of V2X CP and I2X CP. As shown in Fig.~\ref{fig:harsh_noise_V2XSet}, all the I2X CP models show better accuracy than the V2X CP models in all of the noises on the V2Xset dataset. Also, the decline rate of the accuracy of I2X CP is lower than that of V2X CP as the noise worsens. Likewise, all the I2X CP models outperform the V2X CP models and have a lower decline rate in V2X-Sim, as shown in Fig.~\ref{fig:harsh_noise_V2X-Sim}.
In addition, the results on V2X-Sim show a significantly large accuracy difference between I2X CP and V2X CP, even in the zero noise. These results indicate that changing the ego agent to infrastructure allows for improved noise robustness because the data of an ego agent is not affected by communication noises; the ego infrastructure utilizes noise-free data itself, thereby enabling I2X CP to show good performance in noisy environments and maintain performance even when the received data worsens.

\subsubsection{Scenario Analysis.}

We compare V2X CP with I2X CP in the same scenarios. For the visualization, we use the detection results of Where2comm~\cite{hu2022where2comm}.

As shown in Tab.~\ref{tab:scenario_accuracy_v2x_i2x}, the accuracy of I2X CP is different depending on the scenarios, and the degree of effectiveness becomes more obvious than Tab.~\ref{tab:scenario_accuracy_v2v_v2x}, which is accuracy comparison between V2V CP and V2X CP. To analyze when I2X CP outperforms V2X CP, we choose two scenarios: Scene \#2 and Scene \#12, which are the most and least effective cases of I2X CP, respectively.

\begin{figure}[t!]
  \centering
  \begin{tabular}{cc}
       \includegraphics[width=0.48\textwidth]{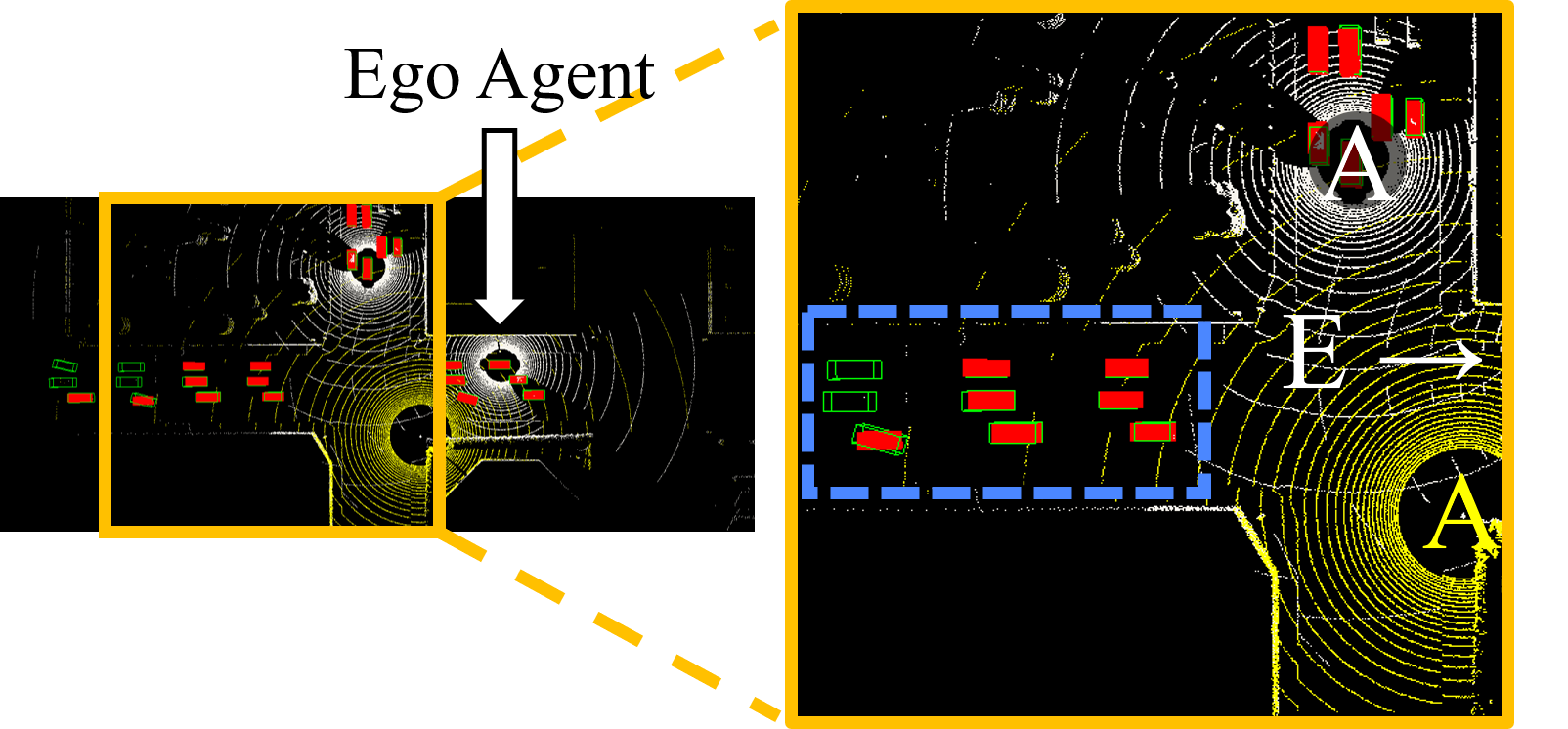} & 
       \includegraphics[width=0.48\textwidth]{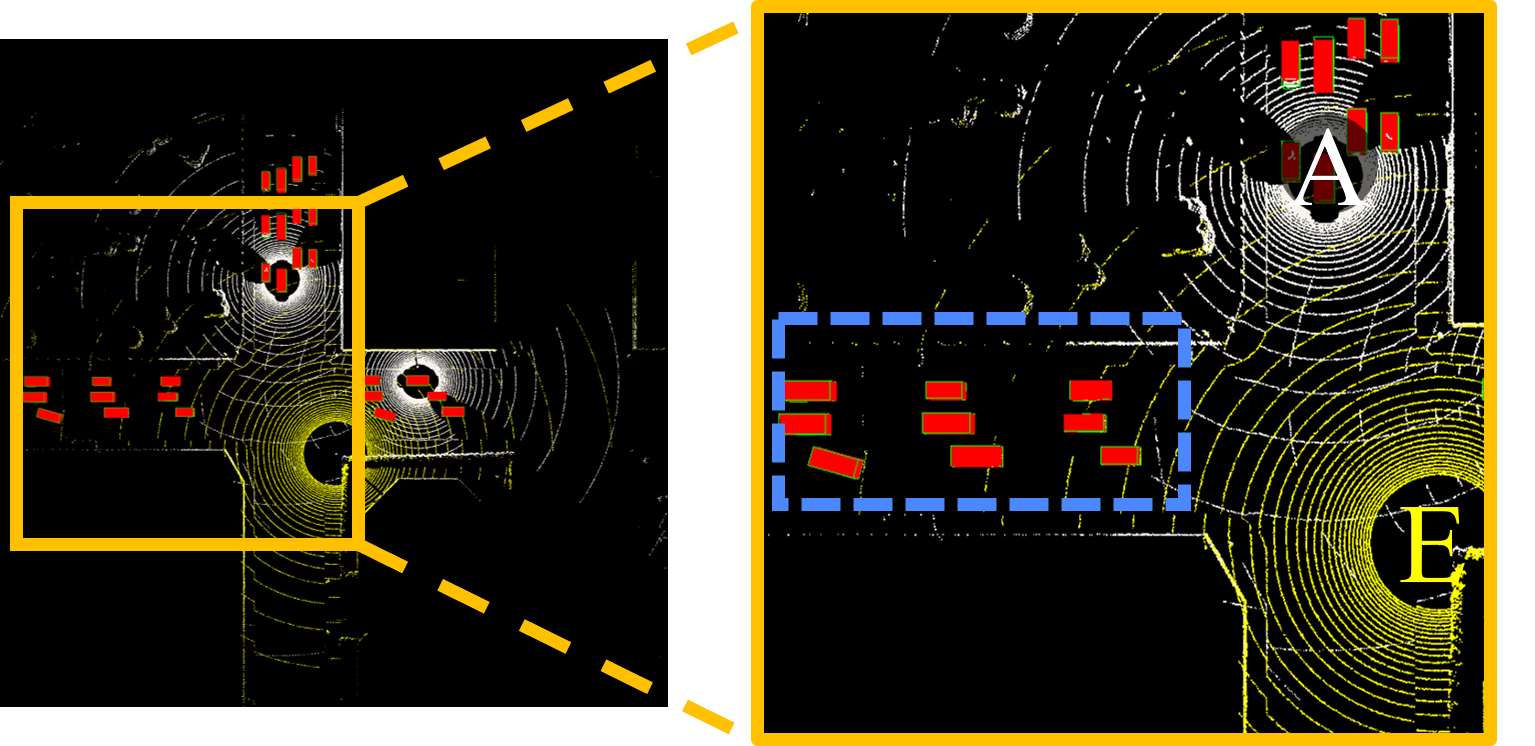} \\
       (a) V2X CP & (b) I2X CP
  \end{tabular}
   \caption{\textbf{Effective case of I2X CP (Scene \#7).} The point clouds from the vehicle's LiDAR are indicated as white dots, and the point clouds from the infrastructure's LiDAR are indicated as yellow dots. The green bounding boxes are ground truth objects, and the red bounding boxes are predicted objects. Both boxes are located in the ego agent's detection range. \textbf{E} means the ego agent and \textbf{A} means aux agents.}
  \label{fig:Scenario_Infra_Centric_Best}
\end{figure}

I2X CP performs better than V2X CP in certain scenarios, such as 4-way and 3-way single intersections. As in the 3-way intersection in Fig.~\ref{fig:Scenario_Infra_Centric_Best}, V2X CP shows errors in detecting distant vehicles when the vehicle heading changes as the ego vehicle turns. In contrast, because infrastructure can monitor all directions equally and does not change heading position, I2X CP shows better detection performance. 

\begin{figure}[t!]
\vspace{10pt}
  \centering
  \begin{tabular}{cc}
       \includegraphics[width=0.48\textwidth]{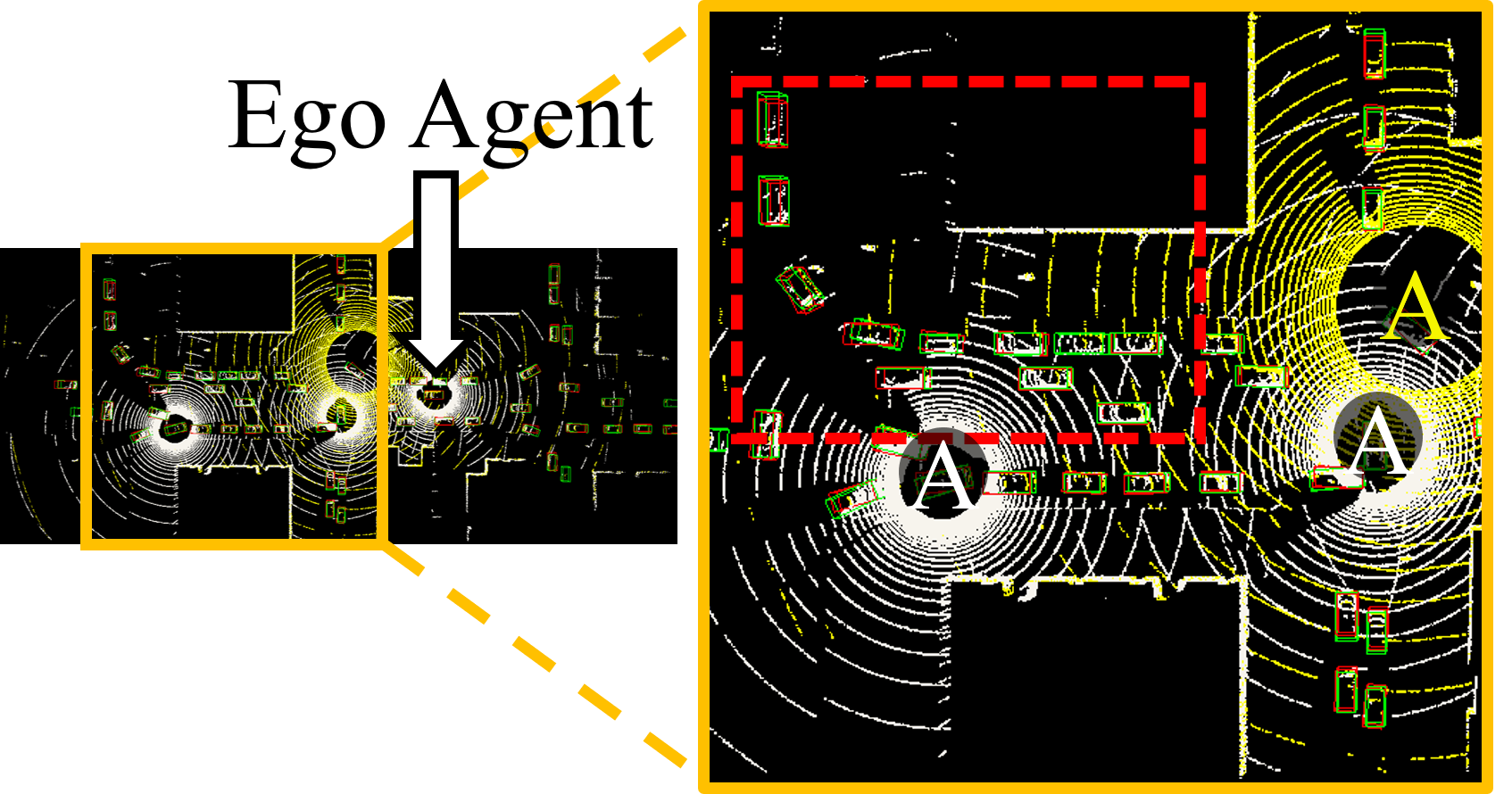} & 
       \includegraphics[width=0.48\textwidth]{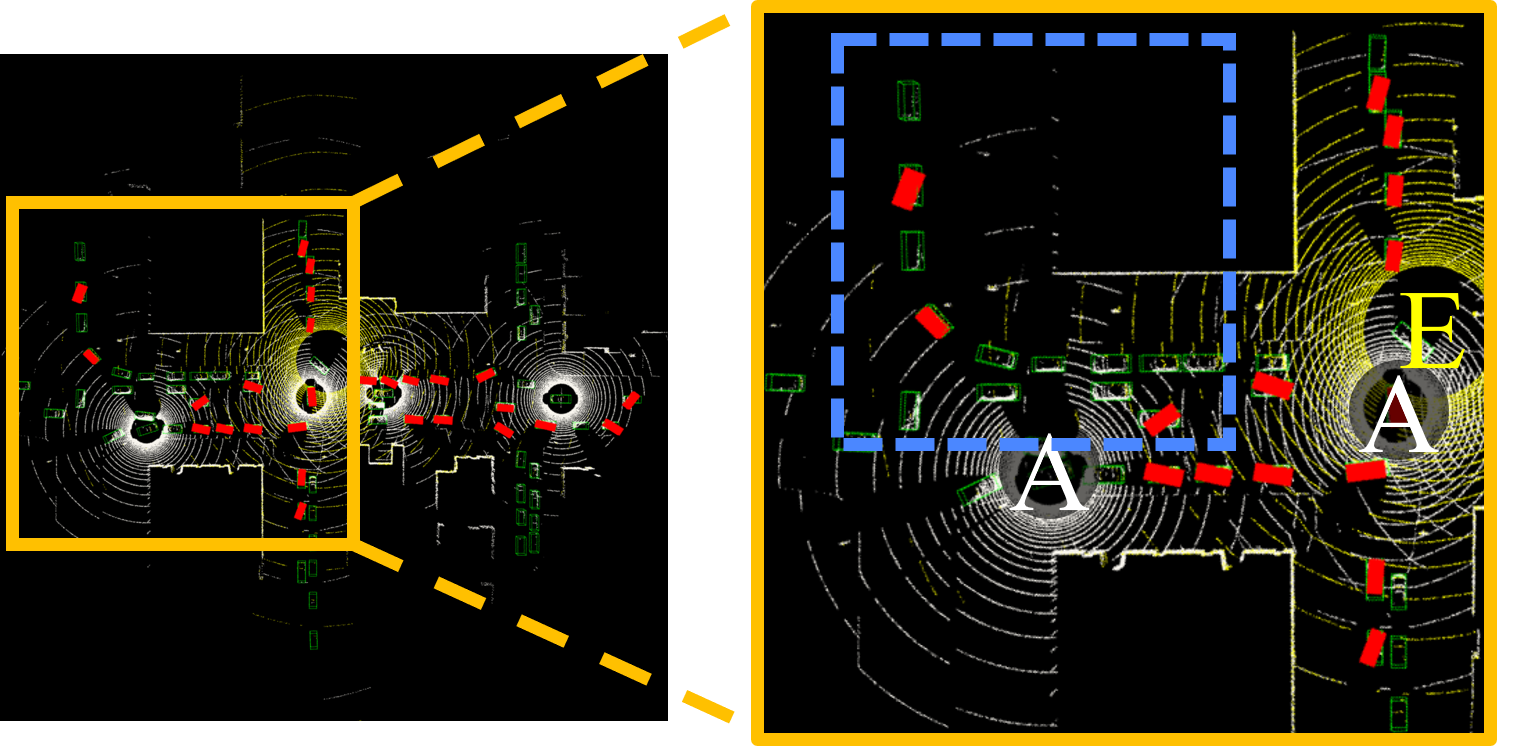} \\
       (a) V2X CP & (b) I2X CP 
  \end{tabular}
   \caption{\textbf{Uneffective case of I2X CP (Scene \#12).} The point clouds from the vehicle's LiDAR are indicated as white dots, and the point clouds from the infrastructure's LiDAR are indicated as yellow dots. The green bounding boxes are ground truth objects, and the red bounding boxes are predicted objects. Both boxes are located in the ego agent's detection range. \textbf{E} means the ego agent and \textbf{A} means aux agents.}
  \label{fig:Scenario_Infra_Centric_Worst}
\end{figure}

On the other hand, V2X CP outperforms I2X CP in certain scenarios, such as two adjacent intersections. As shown in Fig.~\ref{fig:Scenario_Infra_Centric_Worst}, I2X CP cannot leverage the information of the next intersection from the aux vehicle, experiencing serious occlusion and thus worsening the performance. In contrast, 
V2X CP can overcome the serious occlusion as it is going toward the next intersection.

\begin{table}[t!]
\centering
\scriptsize
\setlength{\tabcolsep}{1pt}
\caption{Comparison of AP@0.7 accuracy between vehicle-centric and infra-centric CP across different scenarios. Difference means the change from AP@0.7 in vehicle-centric CP to AP@0.7 in infra-centric CP.}
\label{tab:scenario_accuracy_v2x_i2x}
\renewcommand{\arraystretch}{1.1}
\begin{tabular}{l|C{1.3cm}|C{1.3cm}|C{1.3cm}|C{1.3cm}|C{1.3cm}|C{1.3cm}|C{1.3cm}}
\hline
\rowcolor[HTML]{D7D5D5} 
\multicolumn{1}{c|}{\cellcolor[HTML]{D7D5D5}Model} & CP       & Scene \#2      & Scene \#8      & Scene \#9      & Scene \#10      & Scene \#11      & Scene \#12 \\ \hline
\multirow{3}{*}{V2X-ViT\cite{xu2022v2xvit}}   
& V2X          &0.813&0.824&0.876&0.765&0.728&0.733\\
& I2X          &0.990&0.476&0.913&0.997&0.134&0.069\\
&Difference&\blue{0.176$\uparrow$}&\red{0.347$\downarrow$}&\blue{0.037$\uparrow$}&\blue{0.233$\uparrow$}&\red{0.595$\downarrow$}&\red{0.664$\downarrow$}\\ \hline
\multirow{3}{*}{Where2comm\cite{hu2022where2comm}}
& V2X          &0.849&0.900&0.862&0.924&0.770&0.786\\
& I2X          &0.993&0.514&0.920&0.998&0.178&0.096\\
&Difference&\blue{0.144$\uparrow$}&\red{0.386$\downarrow$}&\blue{0.059$\uparrow$}&\blue{0.074$\uparrow$}&\red{0.593$\downarrow$}&\red{0.690$\downarrow$}\\ \hline
\multirow{3}{*}{ParCon\cite{bae2024parcon}}   
& V2X          &0.870&0.920&0.854&0.833&0.824&0.840\\
& I2X          &0.990&0.445&0.944&0.997&0.198&0.116\\
&Difference&\blue{0.119$\uparrow$}&\red{0.475$\downarrow$}&\blue{0.090$\uparrow$}&\blue{0.164$\uparrow$}&\red{0.626$\downarrow$}&\red{0.725$\downarrow$}\\  \hline
\end{tabular}
\vspace{-5pt}
\end{table}

\section{Conclusion}\label{sec:conclusion}
We have re-examined the role of infrastructure within collaborative perception frameworks, traditionally dominated by vehicle-centric models.
Our study has quantitatively demonstrated that integrating infrastructure data into vehicle-centric CP enhances detection accuracy, particularly in complex environments with occlusions. We have also evaluated infra-centric CP, which shows superior performance in noise robustness and detection accuracy in structured environments, such as intersections. These findings suggest that the optimal CP strategy is context-dependent, varying with specific operational environments and physical characteristics.

We highlight the need for a more nuanced approach to CP that fully leverages the strengths of both vehicles and infrastructure. By redefining the role of infrastructure from a simple auxiliary agent to a potential primary agent, we open up a new, promising direction in collaborative perception. 

\subsubsection{Limitation and Future Work.}
One limitation of infra-centric CP is that its performance largely depends on road scenarios. For example, when the ego agent perceives a wide range, such as two or more intersections, the environment provokes the accumulated error of detecting objects that are located out of sensor range. This problem can be mitigated if the dataset involves data from multiple infrasturctures. We plan to expand infra-centric CP to consider Infra-to-Infra (I2I) communication to overcome the limitation.

\section*{Acknowledgement.}
This research was supported by the MSIT(Ministry of Science and ICT), Korea, under the ITRC (Information Technology Research Center) support program (IITP-2024-RS-2023-00259991) supervised by the IITP(Institute for Information \& Communications Technology Planning \& Evaluation), BK21 FOUR(Connected AI Education \& Research Program for Industry and Society Innovation, KAIST EE, No. 4120200113769), and the Korea Agency for Infrastructure Technology Advancement (KAIA) grant funded by the Ministry of Land, Infrastructure and Transport (Grant 22AMDP-C161754-02).


%
%
\bibliographystyle{splncs04}
\bibliography{main.bbl}
\end{document}